\documentclass[sigconf]{acmart}

\usepackage{xcolor}
\usepackage{multirow}
\usepackage{enumitem}
\usepackage{threeparttable}

\AtBeginDocument{%
  \providecommand\BibTeX{{%
    \normalfont B\kern-0.5em{\scshape i\kern-0.25em b}\kern-0.8em\TeX}}}

\setcopyright{acmcopyright}
\copyrightyear{2021}
\acmYear{2021}
\setcopyright{acmcopyright}\acmConference[CIKM '21]{Proceedings of the 30th ACM
International Conference on Information and Knowledge Management}{November
1--5, 2021}{Virtual Event, QLD, Australia}
\acmBooktitle{Proceedings of the 30th ACM International Conference on Information
and Knowledge Management (CIKM '21), November 1--5, 2021, Virtual Event, QLD,
Australia}
\acmPrice{15.00}
\acmDOI{10.1145/3459637.3482076}
\acmISBN{978-1-4503-8446-9/21/11}

\begin{document}
\fancyhead{}

\title{CoSEM: Contextual and Semantic Embedding for App Usage Prediction}

\author{Yonchanok Khaokaew}

\affiliation{%
  \institution{RMIT University}
  \city{Melbourne}
  \state{VIC}
  \postcode{3000}
  \country{Australia}
}\email{s3771152@student.rmit.edu.au}

\author{Mohammad Saiedur Rahaman}
\orcid{0003-2320-0112}
\affiliation{%
  \institution{RMIT University}
  \city{Melbourne}
  \state{VIC}
  \postcode{3000}
  \country{Australia}
}\email{saiedur.rahaman@rmit.edu.au}

\author{Ryen W. White}
\affiliation{%
  \institution{Microsoft Research AI}
  \city{Redmond}
  \state{WA}
  \country{United States}
}\email{ryenw@microsoft.com}

\author{Flora D. Salim}
\orcid{0002-1237-1664}
\affiliation{%
  \institution{RMIT University}
  \city{Melbourne}
  \state{VIC}
  \postcode{3000}
  \country{Australia}
}\email{flora.salim@rmit.edu.au}

\renewcommand{\shortauthors}{Trovato and Tobin, et al.}

\begin{abstract}
App usage prediction is important for smartphone system optimization to enhance user experience. Existing modeling approaches utilize historical app usage logs along with a wide range of semantic information to predict the app usage; however, they are only effective in certain scenarios and cannot be generalized across different situations. This paper address this problem by developing a model called \textit{Contextual and Semantic Embedding model for App Usage Prediction (CoSEM)} for app usage prediction that leverages integration of 1) semantic information embedding and 2) contextual information embedding based on historical app usage of individuals. Extensive experiments show that the combination of semantic information and history app usage information enables our model to outperform the baselines on three real-world datasets, achieving an MRR score over 0.55,0.57,0.86 and Hit rate scores of more than 0.71, 0.75, and 0.95, respectively.

\end{abstract}

\ccsdesc[500]{Information systems~Decision support systems}
\ccsdesc[500]{Human-centered computing~Ubiquitous and mobile computing}

\keywords{App usage prediction, semantic embedding, profile embedding}


\maketitle

\section{Introduction}

Smartphone applications play a vital role in our daily lives for many reasons, such as social networking, messaging, shopping, and gaming, are to name a few. With the potential of mobile devices, the number of mobile applications (abbr. apps) has grown exponentially in the worldwide market. In particular, there are over 1 million apps in Google Play Store, and an average user installs about 50 on their phones \cite{CAO20171}. Nevertheless, people now use a variety of applications to accomplish a wide range of their daily activities. Therefore, app usage prediction, i.e., which set of apps will be used next by users, is very important for improved user experience and satisfying user information needs at one go. Moreover, app usage prediction can enable efficient system resource management and power consumption optimization.

However, the prediction of app usage is challenging as it depends on different semantics such as tasks, search query, location, and time. For example, suppose a user wishes to edit a document in the future. In that case, an intelligent smartphone assistant can preload the required apps by understanding the semantics (i.e., this is an editing task for a construction engineering client). Several works have attempted to address this challenge in recent years as a next used app prediction problem~\cite{Huang_ubicome2012,Shin_Ubicomp2012,Baeza-Yates.yahoo.2015.WSDM,appusage2vec,Liao_2013_ICDM,DO201479,apprushSUN2013445,hinextapp8029515,Bayesian.HMM.10.1145/2629504, Natarajan_2013, Xu_appbag_ISWC2013, parate_ubicom_appm_2013}. Recent research utilizes Spatio-temporal information such as location, time, and session data to predict the next app that an individual user will use next \cite{Baeza-Yates.yahoo.2015.WSDM}. Another study utilized a Bayesian network on the Spatio-temporal app usage data for predicting the next app~\cite{Huang_ubicome2012, Wang2019_ubicom_appusage}. At the same time, some works aimed to use similar information to predict app usage in different scenarios (e.g., in different places \cite{yu2018smartphone, Fan_2019_transfer_learning} or apps used to complete various search tasks \cite{Aliannejadi.CIKM201810.1145/3269206.3271679}).

Although many methodologies have been proposed to unravel the app prediction problem, they usually require extensive feature extraction \cite{Liao_2013_ICDM}, or context-triggered feature generation (e.g., Yahoo's Aviate application) \cite{Baeza-Yates.yahoo.2015.WSDM} which is challenging without proper domain knowledge. Moreover, these proposed models are usually only applicable to a particular type of dataset (contextual scenario); for instance, the proposed model in \citeauthor{yu2018smartphone} \cite{yu2018smartphone} only can be employed in Points-of-Interest datasets. Since the semantic information for each of these research is different, the adaptation of the current approaches in a generalized manner across all the scenarios cannot be achieved without compromising the prediction performance. In this research, we conduct our investigation by forming the following research question: 
\emph{"Can we address the app usage prediction problem generically, in the presence of various semantic information, without compromising the prediction performance?"}

Then we present a generic approach called Contextual and Semantic Embedding (CoSEM) that can handle various semantic information, such as task information or a search query, and utilize this information and historical app usage data to enhance the performance app usage prediction. CoSEM leverages the latent semantic information, which might be different across datasets, to predict applications used by a user in future timestamps. Rather than using the user's identification to create user profiles \cite{appusage2vec}, we produce the user vector using the user's previously used app, which is contextual knowledge about the user, and then incorporate the semantic information to improve the Dual-DNN for app usage prediction problem. In addition, we are the first to evaluate the app usage prediction problem on three real-world app usage datasets with varying prediction tasks (i.e., CPS-Task~\cite{liono2019building}, ISTAS~\cite{Aliannejadi.CIKM201810.1145/3269206.3271679}, and China Telecom~\cite{yu2018smartphone}). 
These datasets are notably different in terms of semantic information such as tasks, search query, and location and time of a user. Our key contributions in this research are as follows:

\begin{itemize}

 \item We introduce a new app usage prediction problem where the application set is predicted in a generalized manner based on available semantic information.
 \item We develop a model called CoSEM that employs two embeddings for a custom Dual-DNN to model the semantic and contextual representations for predicting a set of apps that a user will use next. An ablation study demonstrates the effectiveness of the joint embeddings.
 \item We demonstrate the generality and effectiveness of our proposed model for app usage prediction with varying semantic information across three real-world datasets: a task and activity dataset, a search behavior dataset, and an app traffic dataset from a telecommunication provider.

\end{itemize}

\section{THE PROPOSED METHOD}

\subsection{Problem Formulation}
 First, we discuss the preliminaries, including semantic and contextual information, to formulate app usage prediction problems.

\textbf{Semantic Information ($\mathcal{S}$) :} 
represents the goal, intent, or condition of users when they use a specific mobile application. The semantic information of a user $u$ in time window $T$ can be defined as {$\mathcal{S}^u_T$} =
\{{$s^u_1 ,s^u_2,...,s^u_m $}\}, where $s$ is the chunk of semantic information (e.g., words in the search query, tasks, location, time), and $m$ is the total number of pieces in semantic information.

\textbf{Contextual information ($\mathcal{A}$) :} used in this work is the user's historical application usage. The contextual information of the user $u$ can be represented as {$\mathcal{A}^u_{t<T}$} = 
\{{$a^u_1 ,a^u_2,...,a^u_n $}\}, where $a^u$ is the application, and $n$ is the number of historical applications used by a particular user $u$ before the time window $T$.

\textbf{App usage prediction}: A model $z$ is developed and trained with $\mathcal{S}$ and $\mathcal{A}$. Specifically, given a semantic information $\mathcal{S}^u_T$ in current time window $T$ and the contextual information $\mathcal{A}^u_{t<T}$, the trained model $z$ predicts a set of applications to be used during $T$. This problem can be formally defined as follows.
\begin{equation}
\label{eqn:problem}
    z(\mathcal{S}^u_T, \mathcal{A}^u_{t<T}) \rightarrow \mathcal{P}^u_T
\end{equation}

Here, $\mathcal{P}^u_T$ is the set of predicted applications.

\subsection{Proposed Model}
\begin{figure}[!h]
  \includegraphics[width=0.75\linewidth,height=6 cm]{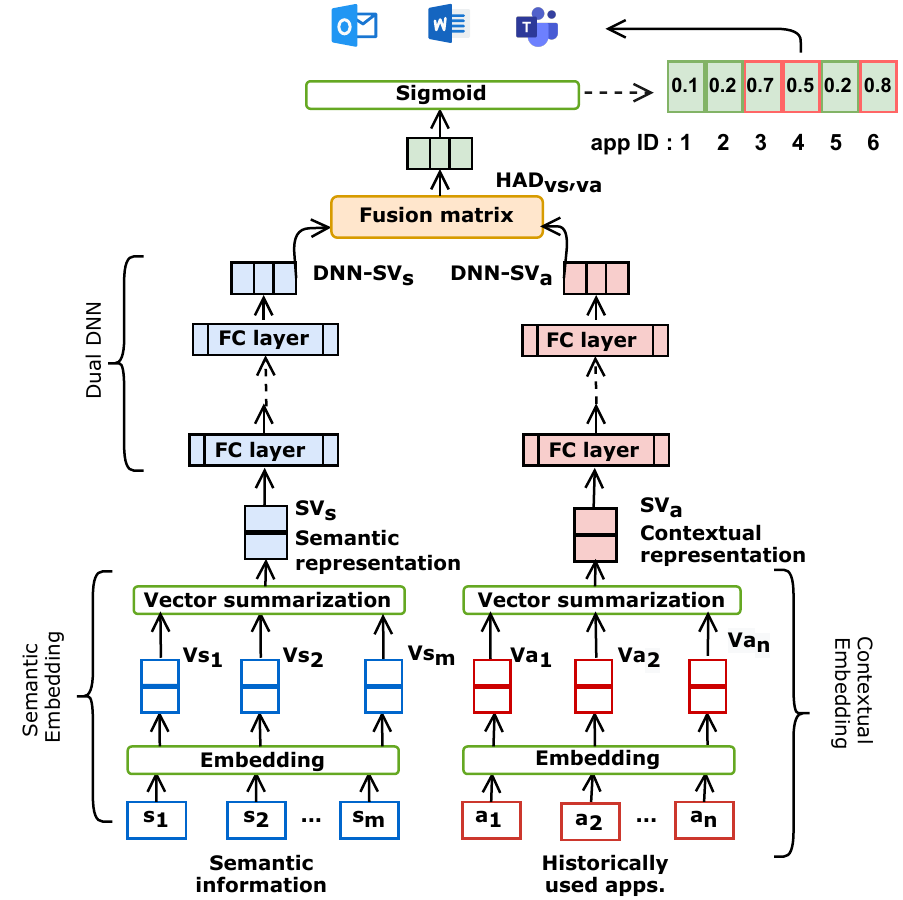}
  \caption{The overall framework of CoSEM}
  \label{fig:flowchart_task_model}
\end{figure}

This section discusses our app usage prediction model called Contextual and Semantic Embedding (CoSEM). The overview of CoSEM is illustrated in figure \ref{fig:flowchart_task_model}. In CoSEM, the chunk of semantic information in the time $T$ will be mapped to a unique vector $v_s$ in matrix $MS$, and an app $a$ in the historical app list, used in the previous period ($t<T$), is also mapped to a vector $v_a$ in matrix $MA$. The $MS$ and $MA$ matrices contain all possible semantic information chunks and applications used in each dataset. CoSEM takes $M$ semantic information vectors, and $N$ used app vectors as input to predict the set of apps used in time $T$. These matrices $MA$ and $MS$ are shared with all users, i.e., the app of ‘Youtube’ vector and the semantic information vector of a word 'recipe' are the same for all users. The values in these matrices will be updated when we train the model.

\subsection{Embedding module}
The first layer is the embedding layer. Through this layer, the chunk of semantic information or historically used apps will map to a low-dimensional dense vector $V_s$ and $V_a$ using an embedding lookup table E $\in$ $R^{Q\times D}$, where $Q$ can be vocabulary size, location size, or possible application size depending on the type of inputs and D is the embedding dimension. The parameters of this embedding layer are tuned during model training.

\textit{Semantic information embedding}
The semantic information in each dataset can be different. This experiment evaluates our framework with three types of semantic information, including the task description, the task query, and the location id and time, from 3 different real-world datasets. To find the representation of semantic information, firstly, we divide the data into chunks and then remove the irrelevant chunks (e.g., for the task description, we remove all stop words contained in the task description). After that, we embed the chunk of semantic information via an embedding layer to vectors $Vs_1$, $Vs_{2}$,...,$Vs_m$ and then feed these vectors as the input to the vector summarizing process based on the following equation $SV_{s} = \frac{\sum_{l=1}^{m}{V_{s_{l}}}}{m}$. Where $SV_s$ is the summarization vector of semantic information, $V_{s_{l}}$ is the vector of each semantic chunk, and $m$ is the length of semantic chunks

\textit{Contextual information embedding}
To overcome the limitation of the user vector when the user's data is inadequate, we introduce the concept of contextual information embedding that uses the historical app usage of an individual. The last $n$ most recent apps that a user used in the time slot before $T$ are taken as an input to the embedding layer for mapping the representation vector of each app. Then, these app vectors will be fed to the vector summarization process as illustrated by the equation $SV_{a} = \frac{\sum_{l=1}^{n}{V_{a_{l}}}}{n}$. Where $SV_a$ is the summarization vector of historical apps, $V_{a_{l}}$ is the vector of app $l$, and $n$ is the length of historical app sequences.

\subsection{Dual-DNN}
After obtaining the representation vectors of semantic information and contextual information, we forward them to the Dual-DNN model from the Appusage2Vec model \cite{appusage2vec} as input features.
In that paper, the Dual-DNN is designed to incorporate user behavior and app usage information. We adapted this Dual-DNN to utilize personalized user behavior regarding app usage and the semantic information for app prediction. It consists of two parallel deep neural networks, while each DNN is the fully connected layers using Tanh as the activation function. In this work, these DNNs will receive the vector from the semantic information embedding process, and the historical app usage vector. Finally, we combine these two vectors using the Hadamard product, which is the point-wise multiplication, instead of using a concatenation operation. The simple concatenation treats the contextual vector and semantic vector separately, and the point-wise multiplication can build the vector representing both semantic and contextual information. The outcome from the Hadamard product will be taken as input in the sigmoid layer to calculate the probability of each app.

\section{Experiment and Results}
We evaluate our proposed model, CoSEM, on the app usage prediction problem based on three types of semantic information, detailed in the following section.

\subsection{Datasets}
We discuss three real-world datasets used in our research to demonstrate the generality and effectiveness of our proposed model for app usage prediction with varying semantic information. 

\subsubsection{CPS-Task Dataset \cite{liono2019building, liono2020intelligent}}
This collection contains anonymous CPS (Cyber, Physical, Social) activity logs corresponding to different user tasks. Tasks in this dataset are annotated by using the Experience Sampling Method \cite{hektner2007experience}
 and the Day Reconstruction Method \cite{czerwinski2004diary}. A total of 53 participants cumulatively recorded 6,321 task instances. In this paper, we do not consider task categories with minimal information ("none" and "other"), and apps rarely used in this dataset (below than ten times), in our experiments.

 \textit{Prediction task.}
We attempt to predict the set of apps used by a user to perform a task in a particular time window with the semantic information and the user's historically used apps.

\textit{Semantic features.} The semantic feature considered in this dataset is a "task description" (ex. "travel from work") since it contains the task condition and objective of the task. These are essential information for the model to select suitable apps. 


%

\subsubsection{ISTAS Dataset \cite{Aliannejadi.CIKM201810.1145/3269206.3271679}}
This dataset is collected from 255 users via the uSearch application. Participants are asked to report the application to perform a specific search task. For example, if a user uses the search query "grocery store" on the google map, the reported query "grocery store" will be corresponding to the "google map" application. In this work, we eliminate all users who have less than five records due to insufficient data for the training process.

\textit{Prediction task.}
We predict an application that users will use to complete a search task in a given time window. We use a semantic feature and the apps used in the last 24 hours as input.

\textit{Semantic features.} 
The search queries are taken as semantic information as it contains the searchers' intents. This intent can help the model specify what apps would suit completing a specific search.

\subsubsection{China Telecom Dataset \cite{yu2018smartphone}}
This is an app usage dataset collected by China Telecom, a major cellular service provider in China. This dataset contains 1,000 unique devices, 2000 applications, and 9850 base stations (i.e., locations) over a week. Note that this is a subset of the publicly available dataset, not the complete dataset used by the original author and this work \cite{appusage2vec}.

\textit{Prediction task.} 
We utilize this dataset to predict the set of applications that a user will use in a time window based on the user's location and time.

\textit{Semantic features.} The spatial and temporal features used in this dataset are considered to be semantic information. In each location and time, there are applications that users usually used, and this set of applications may vary across the user cohort. For example, semantic information in this dataset is the base station "175" and "4", the time in terms of an hour of the day.

\subsection{Experimental Settings}

\textit{Evaluation metrics.} Efficiency is measured by the hit rate (HR) and the mean reciprocal rank (MRR). We include the MRR score instead of using only the hit rate score because the hit rate score calculates only the hit ratio of the top-k Apps without considering the order of predicted apps. As long as correct apps are in top-k predict apps, it is counted as one hit. In contrast, the MRR considers the order of predicted apps. Therefore, we choose both the MRR and HR as evaluation metrics. Several prior works have used these matrices to evaluate their model's performance \cite{Aliannejadi.CIKM201810.1145/3269206.3271679,NTAS,Liao_2013_ICDM,appusage2vec,Loai2,trust}. We use the following equation ${MRR} = \frac{1}{|Q|} \sum_{i=1}^{|Q|} \frac{ 1 }{ f_{id} }$ to calculate the MRR score. Here, $|Q|$ is the number of testing instances, $f_{id}$ is the first position of correctly predicted apps in each instance. In this paper, we set the number of predicted apps to 5, so the MRR metric will be calculated with these five predicted applications, $f_{id}\in [1, 5]$.

\textit{Data separation.} To evaluate our proposed model, we sort the data of each user in each dataset chronologically and take the first 70\% of each user for the training model, the next 10\% for validation, and the rest for testing. This dataset setting will be applied to all baselines. In the tuning process, the MRR value on the validation set is used to select the hyperparameters for our model.

\subsection{Baselines}

We compare our work to several state-of-the-art models, showing
the performance of our proposed model for the app usage prediction problem from semantic information.

\begin{itemize} [leftmargin=0.31cm]
\item[] \textbf{Most recently used (MRU)} considers the $k$ apps that were the most recently used by each user as predicted results. Note that for ISTAS data, we consider the last applications used for searching the query as the most recently used app.

\item[] { \textbf{NTAS2}} is the Neural Target Apps Selection model presented in \cite{NTAS}. We use the NTAS2 model to compare the performance of our work with the semantic information from each dataset.

\item[] { \textbf{CNTAS}} is the Context-Aware Neural target apps selection, in \cite{Aliannejadi.CIKM201810.1145/3269206.3271679}. This model is the improved version of NTAS, which considers contextual information when selecting the app as a predicted outcome. The CNTAS model creates the representation vector of candidate apps, semantic, and contextual information. The model is trained to ensure that the vector of the right candidate app more similar to the vector of semantic and contextual information than the wrong candidate app (negative app). This makes the CNTAS model different from our work, although we use both the semantic and contextual information. As CPS-Task and China Telecom dataset, we use the previous hour's app usage as the contextual information because the last 24 hours' app usage is not available in those datasets. Note that we use all possible apps as candidate apps, as the proposed paper \cite{Aliannejadi.CIKM201810.1145/3269206.3271679} does not mention the number of candidate apps used in the training process.

\item[] { \textbf{Appusage2Vec}} \cite{appusage2vec} embeds apps and users as vectors and treats them as an input for predicting the next app. The experiment with Appusage2Vec is presented only for the CPS-Task and China Telecom datasets, as the ISTAS dataset lacks the time spent on each app, which is a necessary feature in their framework. 
\end{itemize}

\subsection{Experiment Results}

We compared CoSEM with the baseline approaches, as illustrated in Table \ref{tab:result data}. Other approaches used the sequence of apps and an embedding dimension in the same size as CoSEM to predict the set of apps. As illustrated in Table \ref{tab:result data}, our CoSEM performs best in all of the three datasets. There is a significant improvement in our model on both MRR and HR metrics. This result indicates that combining the semantic information and historically used apps is helpful for the app prediction problem. Compared to the other baseline model, our CoSEM outperforms the Appusage2Vec in both the CPS-task dataset and the China telecom dataset around 0.09 and 0.34 on MRR score, respectively. A similar result compared with the CNTAS and NTAS2 models, our model also receives a better MRR and HR score in all datasets and achieves a 0.1 - 0.2 improvement over these models in terms of HR score. The reproduced result of the CNTAS model is different from their proposed paper because we drop users who have insufficient data, and the number of negative apps is used in the training process. The reason that CoSEM outperforms other baselines due to three reasons. Firstly, Appusage2Vec can achieve the best performance when applied to a vast dataset (more than 8k users, around 315 app sequences/ user), while the CPS-Task dataset only consists of 53, and the China telecom dataset is consists of 1000 users. Second, combining a user vector from the user id and app sequence vector is not enough for predicting the set of apps used to perform the task. Finally, Appusage2Vec, NTAS2, and CNTAS are designed for selecting the only app that will use by the user, not predicting the set of apps used.

\begin{table}
    \footnotesize
    \begin{threeparttable}
  \caption{Performance comparison (*M=MRR,H=Hit ratio)}
  \label{tab:result data}
  \begin{tabular}{lcccccc}
    \toprule
    Dataset & \multicolumn{2}{c}{CPS-Task}& \multicolumn{2}{c}{ISTAS}&
    \multicolumn{2}{c}{China Telecom}\\
    \midrule
    Model/Metric & M@5  & H@5 &M@5  & H@5& M@5  & H@5\\
    \midrule

MRU& 0.2054 & 0.4180& 0.5108 & 0.6383& 0.2168 & 0.3297   \\

NTAS 2 & 0.3406 & 0.8078   & 0.4533 & 0.6925 & 0.4358 & 0.5808   \\
CNTAS & 0.4312 & 0.6029 &  0.3765 & 0.6587 & 0.1797 & 0.4830  \\

Appusage2Vec& 0.4004 & 0.9247  & \multicolumn{2}{c}{N/A}  & 0.4772 & 0.7145   \\

\midrule

DNN-A& 0.8345 & 0.943  & 
0.5644 & 0.7487 & 
0.5509 & 0.7021 \\

DNN-S &0.6747 & 0.8858 & 
0.4527 & 0.7037 & 
0.4349 & 0.5779  \\
 
\midrule

\textbf{CoSEM}&\textbf{0.8681} &\textbf{ 0.9583} & 
\textbf{0.5717} & \textbf{0.7513} &\textbf{0.5571} & \textbf{0.7195} \\

  \bottomrule
\end{tabular}
\begin{tablenotes}
      \footnotesize
\item [] \textit{significance test with p value <= 0.05}
\end{tablenotes}
\end{threeparttable}

\end{table}

\subsection{Ablation Study}
To investigate the effectiveness of CoSEM, we perform ablation study with the following variants of the proposed model:
\begin{itemize}[leftmargin=0.31cm]

\item[]{ \textbf{DNN-A}} \cite{DNN} stacks fully-connected neural layers to learn nonlinear, hidden, and implicit features from a historical app list and then applies logistic regression with a sigmoid layer for predicting an app set. We use the embedding layer to vectorize the historical app list and then input it to fully-connected layers.

 \item[]{ \textbf{DNN-S}} stacks fully-connected neural layers with the embedding layer to embed the information similar to DNN-A, except taking semantic information as input.

\end{itemize}

We found that the ablation of the different parts of our model influenced the average MRR score. The removal of semantic information caused a reduction in both MRR and HR scores which are 0.01 and 0.03 respectively. While the ablation of app usage information resulted in a significant decline in both scores, this indicates that the app usage information is the essential part of our model; however, the model needs the semantic information and app usage information to get the best performance.

\section{CONCLUSION}
In this paper, we proposed a semantic information-aware model called CoSEM for app usage prediction. CoSEM consists of two main parts: including semantic embedding and contextual embedding. In our model, we managed to utilize the historical app usage as contextual information from users and leverage the semantic information to learn the relationship between the semantic information and the set of applications used. 
The experimental results obtained in this study suggest that the semantic information and contextual information based on app usage can jointly create the model for the app usage prediction.
CoSEM can cope with the three different semantic information and achieve noticeable improvement over the baseline methods on three real-world datasets. The MRR and HR scores achieved by CoSEM for predicting the set of applications ranges between 0.55 - 0.86 and 0.71 - 0.95, respectively. In contrast, the best baseline achieves the MRR score between 0.43 - 0.47 and the HR score between 0.69 - 0.92 across the three datasets.

\section*{Acknowledgement}
This research was partially supported by Microsoft Research through Microsoft-RMIT Cortana Intelligence Institute. Any opinions, findings, and conclusions expressed in this paper are those of the authors and do not necessarily reflect those of the sponsors.

\balance
\bibliographystyle{ACM-Reference-Format}
\bibliography{reference}


\end{document}